\documentclass[wcp]{jmlr}

\usepackage{longtable}

\usepackage{booktabs}
\usepackage[load-configurations=version-1]{siunitx} 

\newcommand{\kk}{{\mathbf{k}}}
\newcommand{\xx}{{\mathbf{x}}}
\newcommand{\yy}{{\mathbf{y}}}
\newcommand{\zz}{{\mathbf{z}}}
\newcommand{\mE}{\mathbf{E}}
\newcommand{\X}{\mathbf{X}}
\newcommand{\mK}{\mathbf{K}}

\newcommand{\me}{\textup{e}}
\newcommand{\figref}[1]{\hyperref[#1]{Figure~\ref{#1}}}
\newcommand{\secref}[1]{\hyperref[#1]{Section~\ref{#1}}}

\theorembodyfont{\upshape}
\theoremheaderfont{\scshape}
\theorempostheader{:}
\theoremsep{\newline}

\jmlrvolume{1}
\jmlryear{2017}
\jmlrworkshop{ICML 2017 AutoML Workshop}

\title[Robust Bayesian Optimization]{Robust Bayesian Optimization with Student-$t$ Likelihood}

\author{\Name{Ruben Martinez-Cantin}\Email{ruben@sigopt.com} \\
  \addr SigOpt Inc.\\ Centro Universitario de la Defensa, Zaragoza
  \AND
  \Name{Michael McCourt} \Email{mccourt@sigopt.com}\\
  \Name{Kevin Tee} \Email{kevin@sigopt.com}\\
   \addr SigOpt Inc.}

\begin{document}

\maketitle

\begin{abstract}
Bayesian optimization has recently attracted the attention of the automatic machine learning community for its excellent results in hyperparameter tuning. BO is characterized by the sample efficiency with which it can optimize expensive black-box functions. The efficiency is achieved in a similar fashion to the \emph{learning to learn} methods: surrogate models (typically in the form of Gaussian processes) \emph{learn} the target function and perform intelligent sampling. This surrogate model can be applied even in the presence of noise; however, as with most regression methods, it is very sensitive to outlier data.  This can result in erroneous predictions and, in the case of BO, biased and inefficient exploration. In this work, we present a GP model that is robust to outliers which uses a Student-$t$ likelihood to segregate outliers and robustly conduct Bayesian optimization. We present numerical results evaluating the proposed method in both artificial functions and real problems.
\end{abstract}

\section{Introduction}
\label{sec:intro} 
Bayesian optimization has become the state-of-the-art for hyperparameter tuning of expensive-to-train systems. The sample efficiency and the black-box approach are the most appealing features of Bayesian optimization to be used in many environments and systems. There are a plethora of studies dealing with performance of machine learning systems by tuning the hyperparameters using Bayesian optimization methods, such as \cite{Snoek2012,HutHooLey11-smac,Bergstra2011}. Following the black-box approach, little knowledge is required of the system to be optimized, allowing even stochastic outcomes. However, most methods assume that the system is well behaved in the sense it does not produce \emph{outliers} or adversarial outcomes. Recently, there has been some work dealing with certain robustness in Bayesian optimization, either from input noise \citep{ubonogueira}, by combining sources from different fidelities \citep{NIPS2016_6118,klein-aistats17,forrester2007multi} or by using MCMC for the regression model \citep{Snoek2012,NIPS2016_6117}. To the author's knowledge, this is the first work on robust Bayesian optimization in the statistical testing sense.

In the context of hyperparameter tuning, there are many situations that can produce unexpected outcomes (outliers). These outliers might appear as a result of random occurrences like a bug in the code, a failure in the system, a database problem, or a network issue. They can also appear from alternate sources as in multi-fidelity environments or from security issues such as an exploited vulernability. Being able to identify and remove outliers is crucial to building a robust system.

Outliers can be devastating for regression, where a single outlier may result in a large estimation error \citep{gelman2014bayesian}. Therefore, outliers might also be problematic for Bayesian optimization because the sampling process relies on a surrogate regression model. For example, an outlier near the optimum might result in the predictions of bad outcomes in the neighborhood, resulting in the area being undersampled or not sampled at all and reducing the possibility of improving the model in future iterations. Furthermore, the fact that outliers are distributed independently of true values may result in numerical issues and stability problems while estimating the parameters of the surrogate model.

In the context of optimization, we can distinguish two kind of outliers: \emph{false positives} (outliers that produce better results than they should) and \emph{false negatives} (outliers that produce worse results). False positive outliers could occur, for example, if, given a particular hyperparameter configuration, a user may mistakenly train on (and overfit to) a small fraction of the data. On the contrary, a false negative outlier could appear as the result of gradient descent run accidentally terminating prior to convergence. While false negatives affect Bayesian optimization indirectly through the surrogate model, false positives might also complicate the identification of the ``optimal'' point. Thus, in many scenarios, false positives might require a perfect detection mechanism to guarantee that the correct optimum is returned.

We present a method for hyperparameter tuning using Bayesian optimization, while simultaneously identifying and removing outliers through a Gaussian process with Student-$t$ likelihood. Our method shows improvement across several benchmarks and applications. Although our method theoretically can deal with false positives and negatives, in the present work we have limited the results to false negatives. Future research is needed to guarantee the identification of the correct optima.

\section{Bayesian Optimization}
\label{sec:bo}
Bayesian optimization refers to a class of primarily black-box optimization strategies that relies on probabilistic surrogate models and decision making to improve sample efficiency. This article follows the most common Bayesian optimization based on a Gaussian process (GP) to define the surrogate model of the function of interest.
Given all previous observations $\yy=\yy_{1:t}$ at points $\X=\X_{1:t}$, where
we assume a Gaussian observation model with homoscedastic noise $y = f(\xx) + \epsilon$, with $\epsilon \sim \mathcal{N}(0, \sigma^2_n)$, the GP model gives predictions at a query point $\xx_q$ at step $t$ which have a normal distribution $y_q \sim \mathcal{N}(\mu(\xx_q), \sigma^2(\xx_q))$ with:
\begin{equation}
  \label{eq:predgp}
  \begin{split}
    \mu(\xx_q) &= \kk(\xx_q,\X)^T\mK^{-1}\yy, \\
    \sigma ^2 (\xx_q) &= k(\xx_q, \xx_q) - \kk(\xx_q,\X)^T \mK^{-1} \kk(\xx_q,\X),    
  \end{split}
\end{equation}
where
\begin{align*}
	\kk(\xx_q,\X) = \begin{pmatrix}k(\xx_q,\xx_1) &\ldots& k(\xx_q,\xx_t)\end{pmatrix}^T\!, \quad
  \mK = \begin{pmatrix}\kk(\xx_1,\X) &\ldots& \kk(\xx_t,\X)\end{pmatrix} + \mathbf{I}\sigma^2_n.
\end{align*}

The kernel is chosen to be the Mat\'ern kernel with $\nu=5/2$, also called $C^4$ Mat\'ern kernel \citep{fasshauer2015kernel},
\begin{equation}
  \label{eq:kernel}
	k(\xx, \xx') = \left(1 + r + r^2 / 3\right) \me^{-r}, \text{ where }r=\|\xx-\xx'\|_{\mE}
\end{equation}
for some positive definite matrix $\mE$.  The automatic relevance determination kernel which we use here restricts $\mE$ to being diagonal. The hyperparameters of $\mE$ are estimated by maximum likelihood, although MCMC methods could also be used \cite{Snoek2012}.

For a initial set of points (5 in the experiments), \emph{latin hypercube sampling} is used. For subsequent points, the acquisition function, by which a new point is chosen, is the expected improvement \citep{Mockus78},
\begin{equation}
  \label{eq:ei}
  EI(\xx) = \mathbb{E}_{p(y|\xx,\mE)} \left[\max(0,y^* - y)\right] = \left(y^* - \mu(\xx)\right) \Phi(\zz) + \sigma(\xx) \phi(\zz),
\end{equation}
where $y^*=\max(y_1,\ldots,y_t)$ is the current incumbent and $\Phi(\cdot)$ and $\phi(\cdot)$ are the CDF and PDF of a normal distribution with $\zz = (\xx - \mu(\xx)) / \sigma(\xx)$.

\section{Robust Bayesian Optimization}
\label{sec:robust}

For GP based Bayesian optimization, the observation model is usually the Gaussian likelihood, allowing closed form inference. However, as shown in \figref{fig:example}, this is very sensitive to outliers. Our approach consists of using a surrogate probabilistic model based on a large tail distribution as the observation model, like the Laplace, the hyperbolic secant, or the Student-$t$ likelihood. All those observation models are robust to the presence of outliers, with the Student-$t$ likelihood usually providing the best results \citep{jylanki2011robust}. \cite{ohagan1979outlier} proved that the Student-$t$ distribution can reject up to $m$ outliers tending to infinity (or negative infinity) provided that there are at least $2m$ observations at all. That article also showed that the Gaussian distribution is \emph{outlier-resistant}, meaning that no outlier will be ever rejected.

The Student-$t$ likelihood model have the form
\begin{equation}
  \label{eq:likt}
t(y;f,\sigma,\nu) = \frac{\Gamma\left(\nu + 1/2\right)}{\Gamma(\nu/2)}\frac{1}
{\sqrt{(\nu\pi)\sigma}}\left[1+\frac{(y-f)^2}{\nu\sigma^2}\right]^{-\nu - 1/2}\!\!\!\!\!\!\!\!\!\!\!\!\!\!\!\!\!\!,
\end{equation}
where $f = f(\xx)$, $\nu$ is the degrees of freedom and $\sigma$ is the scale parameter.
However, the Student-$t$ likelihood, as well as the alternative distributions mentioned, do not allow closed form inference of the posterior. \cite{vanhatalo2009student} first suggested to use the Laplace approximation the compute the posterior inference. The same authors later compared different strategies in \cite{jylanki2011robust}, that is, the MCMC from \cite{neal1997monte}, variational approaches and expectation propagation (EP). They showed that a modification of the EP is the most robust estimation method, although with an increased computational cost. We found that, in the context of Bayesian optimization where observations arrive sequentially, the Laplace approximation works fine. 

The Laplace approximation for the conditional posterior $p(f | \yy, \X, \mE, \sigma, \nu)$ of the latent function is constructed from the second order Taylor expansion of log posterior around the mode $\hat{f}$, which results in a Gaussian approximation:
\begin{equation}
  \label{eq:postst}
  p(f | \yy, \X, \mE, \sigma, \nu) \approx \mathcal{N}(f | \hat{f}, \Sigma),
\end{equation}
where $\hat{f} = \arg\max p(f | \yy, \X, \mE, \sigma, \nu)$ is the
maximum \textit{a posteriori} and $\Sigma^{-1} = \mK^{-1} + \mathbf{W}$ the
Hessian of the negative log conditional posterior at the mode with,
\[
\mathbf{W} = diag_i\left(\nabla_{f_i} \nabla_{f_i} \log p(y|f_i, \sigma, \nu)|_{f_i=\hat{f_i}}\right).
\]
For the computation of the Laplace approximation, we have used the \cite{gpy2014} library. We refer to \cite{vanhatalo2009student} for implementation details about the posterior inference.

\begin{figure}
	\floatconts
	{fig:example}
	{\caption{Regression with outliers using the Gaussian likelihood can yield biased estimates and high variance \emph{(left)}. The Student-$t$ likelihood process allows for a better regression, but the estimate with respect to the non-corrupted values is biased and numerically instable \emph{(center)}. We use the Student-$t$ likelihood to remove the outliers and use a Gaussian likelihood with the remaining points \emph{(right)}.}}
	{
		\includegraphics[width=0.3\linewidth]{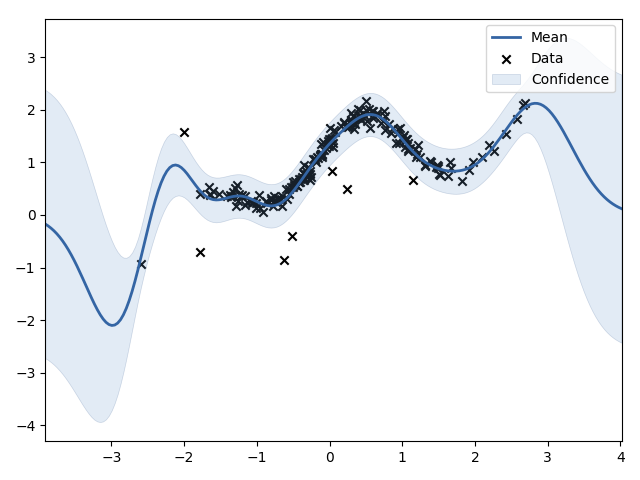}
		\includegraphics[width=0.3\linewidth]{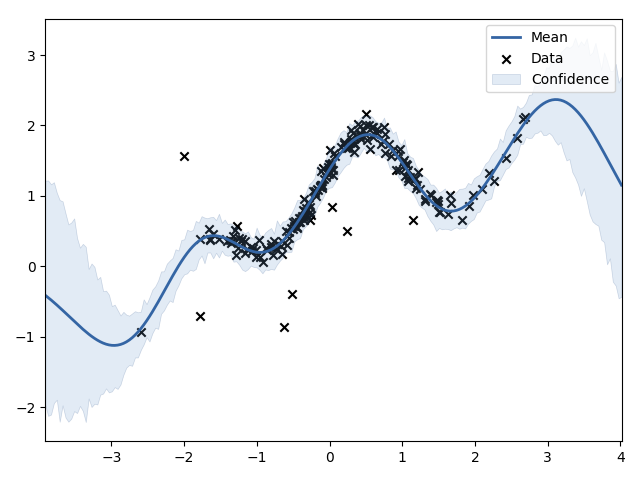}
		\includegraphics[width=0.3\linewidth]{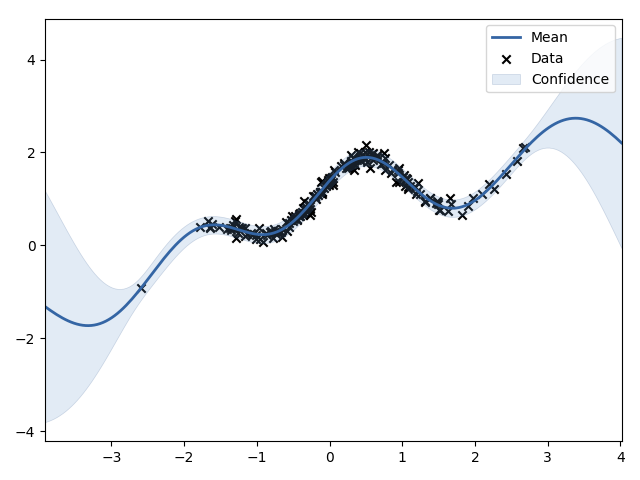}
	}
\end{figure}

Once we have built the regression model with the Student-$t$ likelihood we are able to identify the outliers from the rest of the data. For that purpose, we compute the upper or lower quantile (1\% or 5\%) as a classification threshold. In theory, assuming that a single observation arrives per iteration, only that last observation should be questioned, but because the model is sequentially improved, we found that reclassifying all the points worked better, as new information allows better classification over past observations. Sometimes, points that initially were considered outliers can be found part of the model while, more frequently, points that were initially misclassified as acceptable are properly detected with a better model.

Although the Student-$t$ likelihood is able to identify $m$ outliers out of $2m$ points, we have found that in practices it is reasonable to wait for a certain number of iterations before starting classifying data. We found that waiting about 20\% or 30\% of the budged works for many scenarios. We also found that, because of the sequential nature of Bayesian optimization, if the last point is misclassified as an outlier, there is a large probability that will be selected again in the next iteration, wasting resources. Furthermore, the computational cost of the Student-$t$ likelihood is much more expensive than the Gaussian likelihood. Therefore, we propose to use the Student-$t$ likelihood only once out of $n$ iterations. Finally, once the outliers are classified and removed, the optimization is performed with a standard GP computed only with the remaining points, because it produces more stable and faster solutions (see \figref{fig:example}).

The use of non-Gaussian observation models has also been used in the past for Bayesian optimization although in the context of preference learning \citep{Brochu07NIPS,Javier-Gonzalez:2017aa}. The Student-$t$ distribution has also been used in the past for Bayesian optimization in the context of Student-$t$ process \citep{O'Hagan1992,martinez2014bayesopt,AmarShah2014}.

\section{Experiments}
\label{sec:experiments}

We evaluated our method on a set of artificial functions and realistic applications. In all cases, the outliers were artificially generated so that the distribution is equivalent for all the methods.  Results are compared between optimization as described in \secref{sec:bo} using the outliers as well as by removing the outliers as described in \secref{sec:robust}; as a baseline, we conduct the optimization without any outliers present in the data.

\begin{figure}
\floatconts
  {fig:matplot}
  {\caption{Optimization of 8D RHKS functions randomly generated. Top: Within model comparison (Mat{\'e}rn kernel). Bottom: Out-of-model comparison (RQ kernel)}}
  {
    \includegraphics[width=0.32\linewidth]{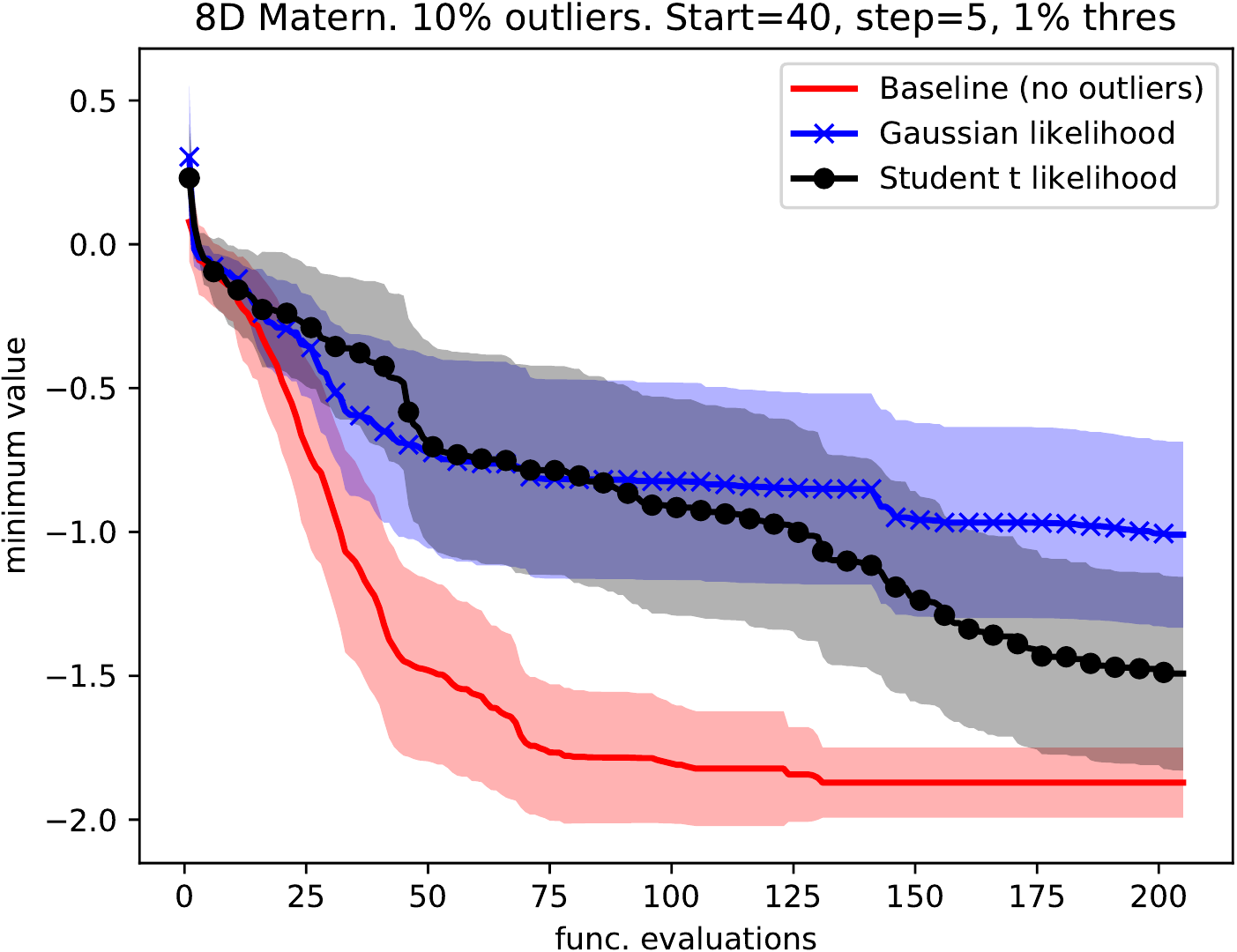}
    \includegraphics[width=0.32\linewidth]{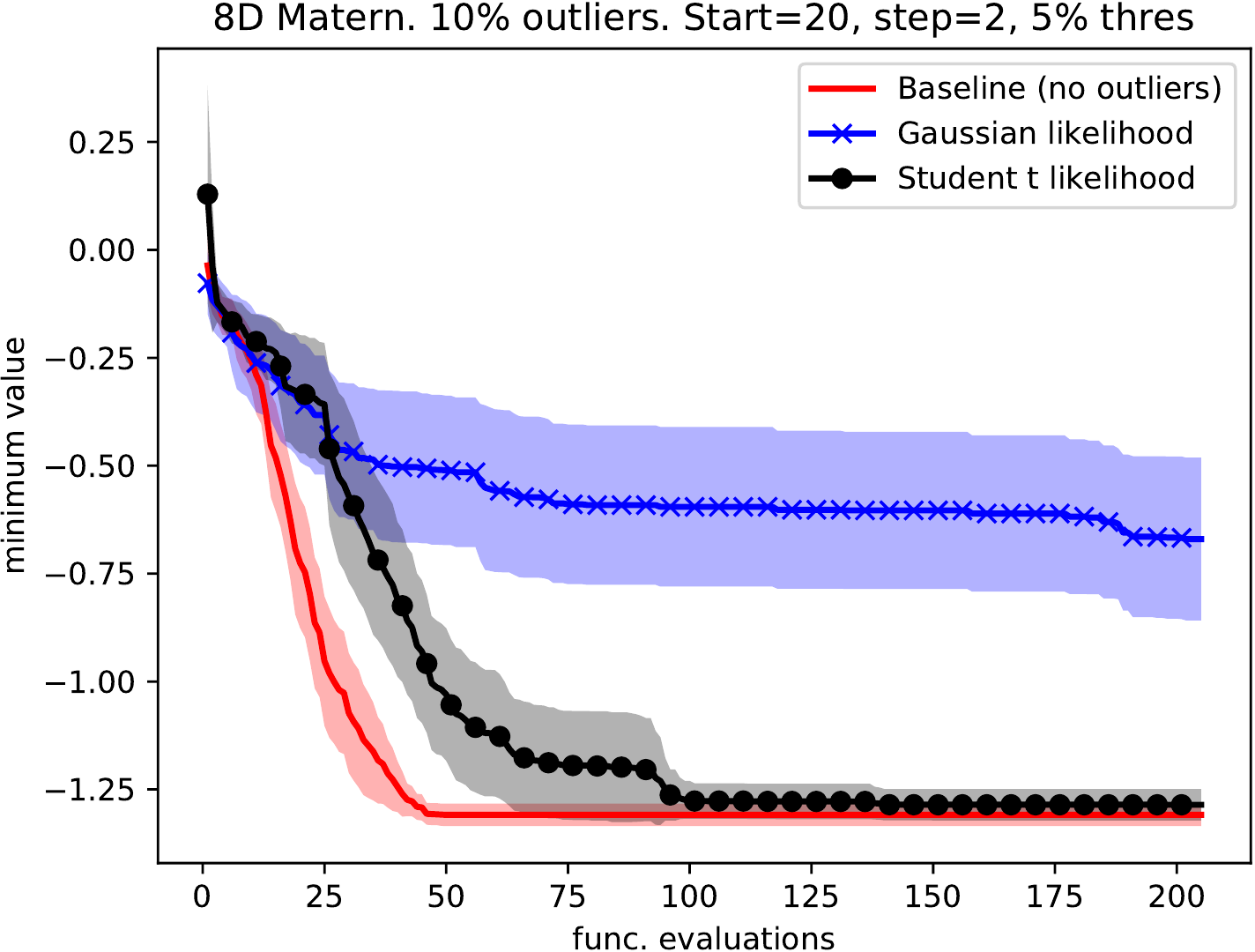}
    \includegraphics[width=0.32\linewidth]{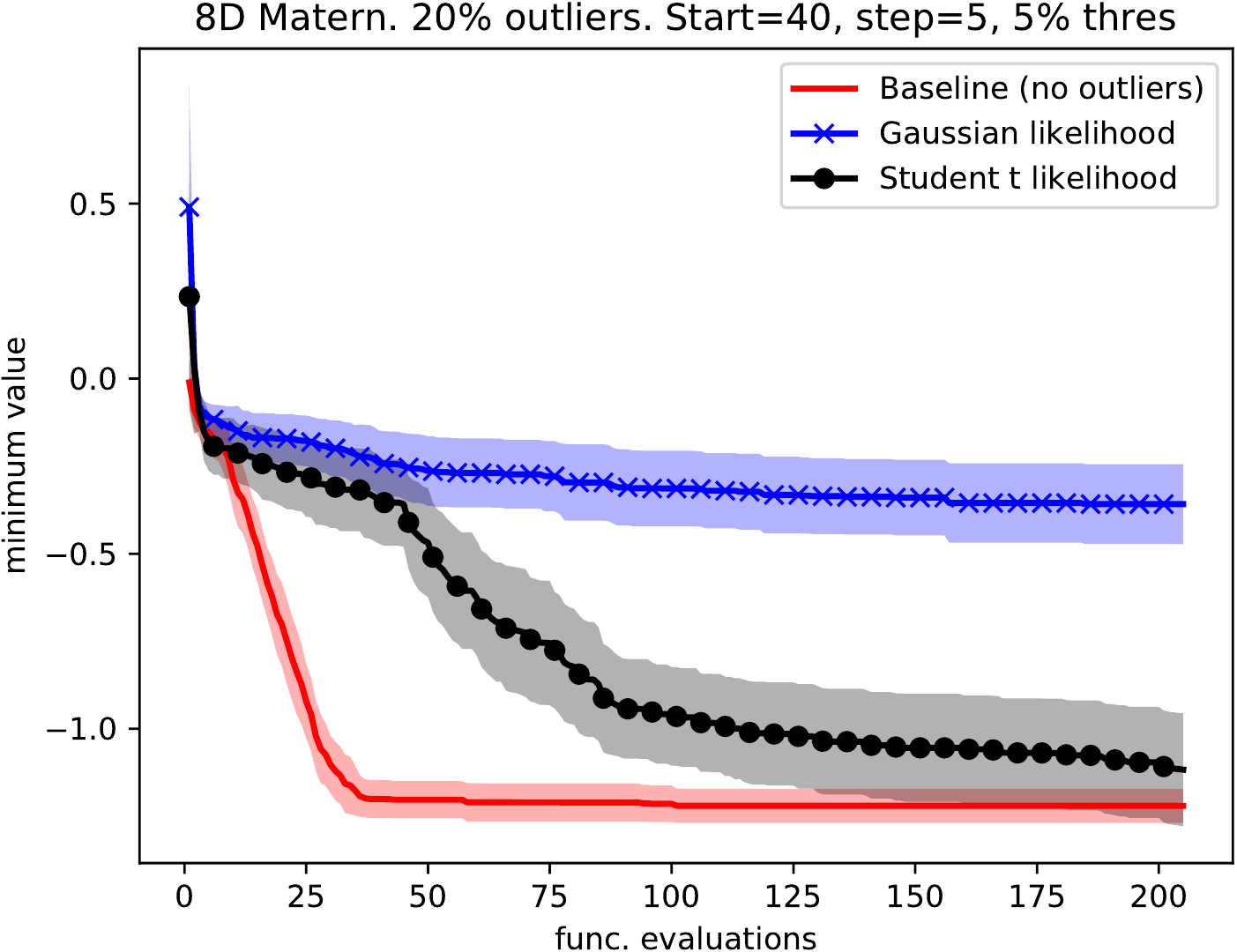}\\
    \vspace{.1cm}
    \includegraphics[width=0.32\linewidth]{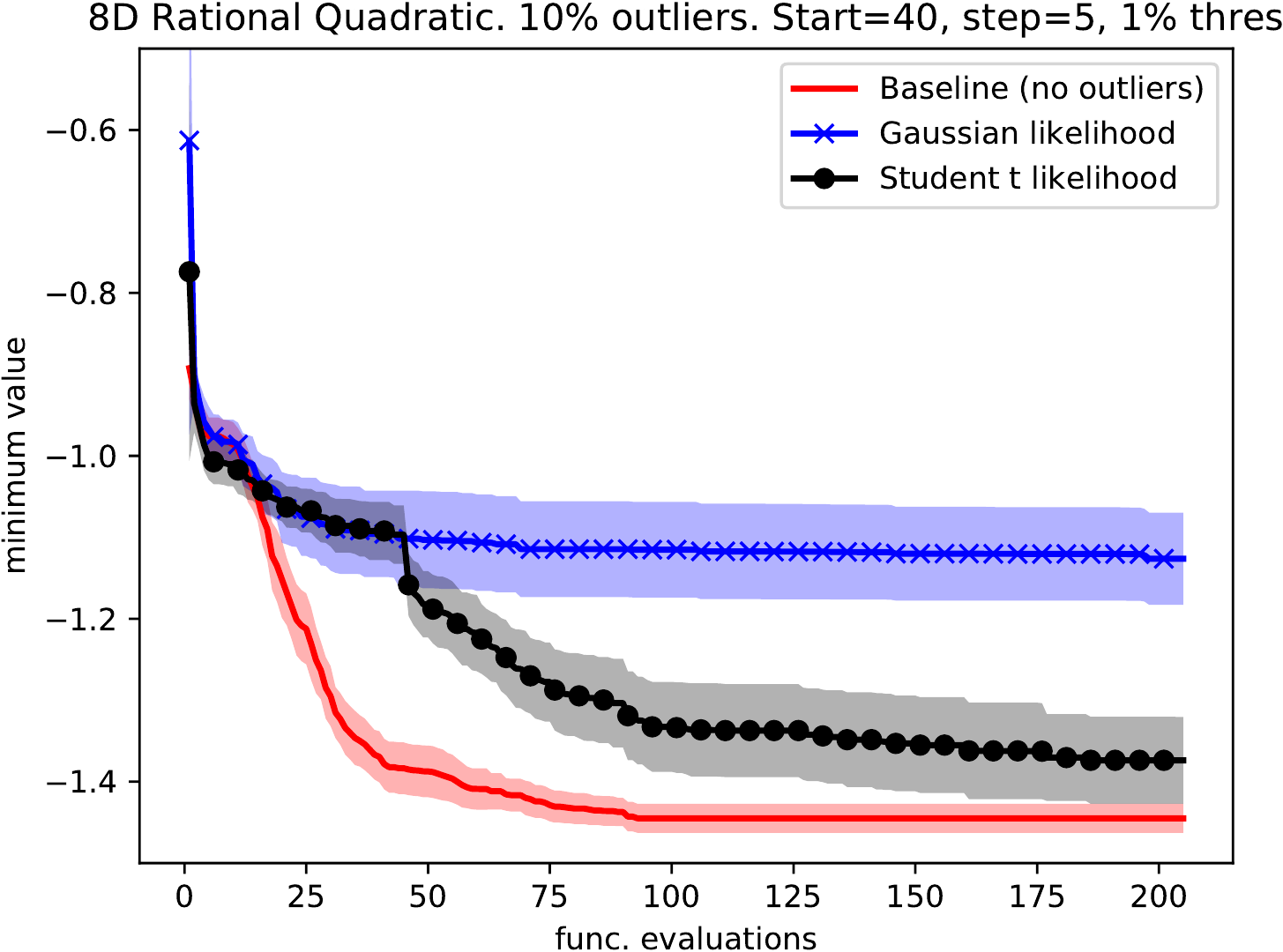}
    \includegraphics[width=0.32\linewidth]{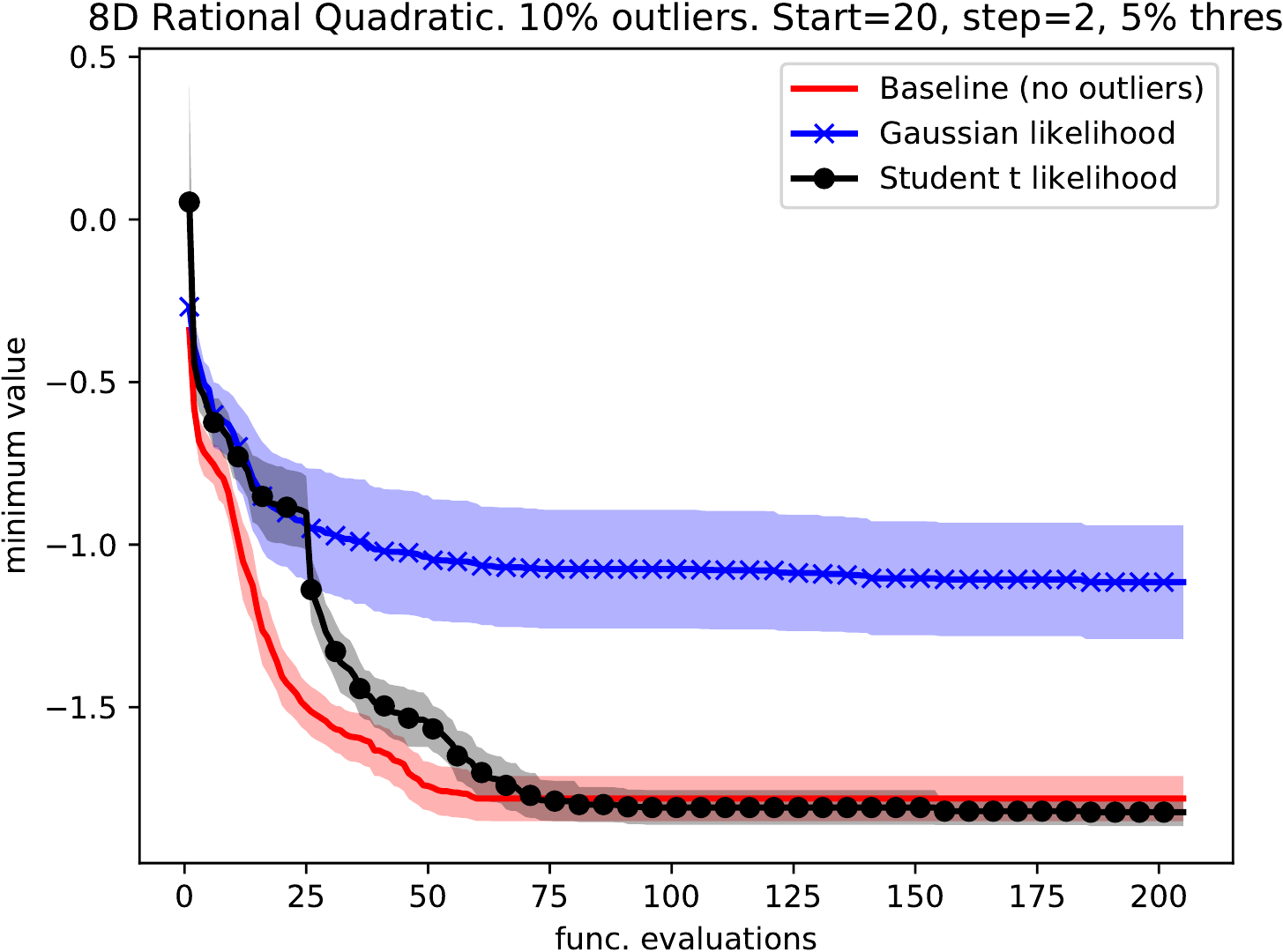}
    \includegraphics[width=0.32\linewidth]{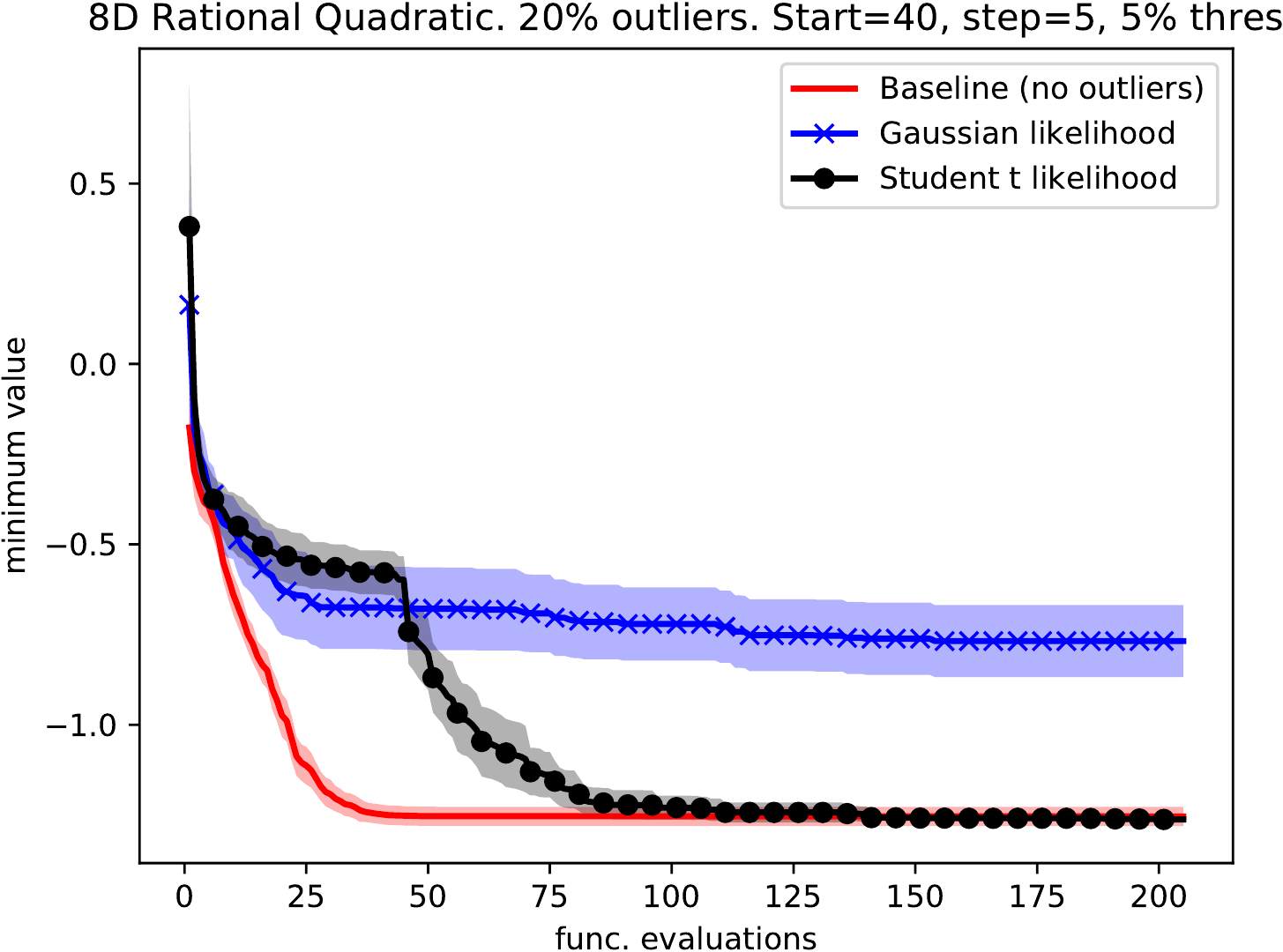}
  }
\end{figure}

\subsection{Numerical benchmarks}

For the numerical benchmarks we have used the same methodology as \cite{HennigSchuler2012}. We have generated a set of random functions from two types of Gaussian process. For the \emph{within model comparison}, we have generated the samples from the same a Gaussian process with the same kernel from \eqref{eq:kernel} as the one being used for the optimization (the top half of \figref{fig:matplot}).

For the \emph{out-of-model} comparison, we have generated the samples from a Gaussian process with a rational quadratic kernel with $\alpha=2$, as shown in the bottom of \figref{fig:matplot}. In both cases, the outliers were sampled iid from a uniform distribution $\mathcal{U}(1,2)$, which roughly corresponds to the top 15\% tail of the GP prior. \figref{fig:matplot} plots the average and 95\% confidence bounds over 20 trials. We selected 8D problems as many hyperparameter tuning problems fall in the 5--10 range.

\subsection{Robust robot control}

Active policy search \citep{MartinezCantin07RSS} is a reinforcement learning method to control a robot or autonomous agent by refining its policy using Bayesian optimization on the reward function. It has been successfully applied for robot walking in controlled environments \citep{Calandra2015a}. In this case, the objective of the policy is to find a stable policy, even in the presence of external perturbations. However, in some trials, the robot might find obstacles or perturbations that physically impossible to overcome. Thus, the robot returns a poor reward even if the policy is good in other conditions.

For the experiments, we have simulated the failures as the robot reaching a \emph{crash} state at a random time during the trajectory. Therefore, the resulting reward is similar to the reward obtained with a bad policy, which also results in a \emph{crash}. \figref{fig:robotplot} plots are the average and 95\% confidence bounds over 30 trials.

It has been shown that robot policy search is a complex problem for Bayesian optimization due to the non-stationary behavior of many reward functions \citep{MartinezCantin17icra}. A large number of outliers yield an underperforming GP model near the optimum because good results and bad results cannot agree to a single stationary function. Thus, the Student-$t$ likelihood also classifies as outliers those bad points that conflict with the good values, resulting in a subtle improvement. However, further research is needed.

\begin{figure}
\floatconts
  {fig:robotplot}
  {\caption{Optimization of the robot walking policy. When the number of outliers is large ($20\%$, left), the Student-t likelihood allows to recover the performance of the baseline. For a small outlier rate ($10\%$, right), the Student-t likelihood is able to prune some of the out-of-model points which allows better refinement.}}
  {
    \includegraphics[width=0.4\linewidth]{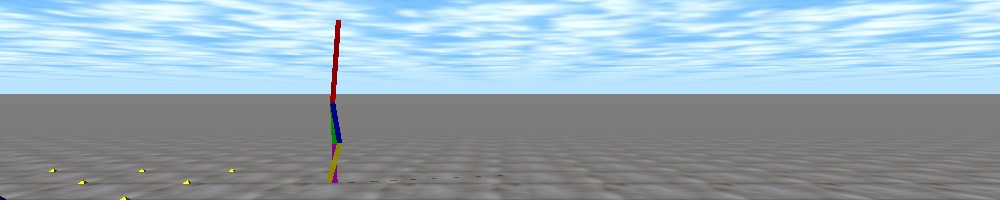}
    \includegraphics[width=0.4\linewidth]{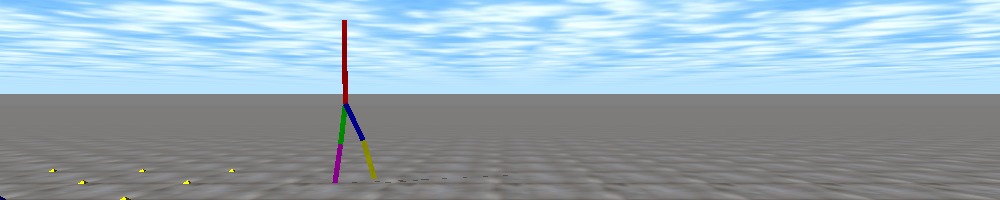}\\
    \includegraphics[width=0.4\linewidth]{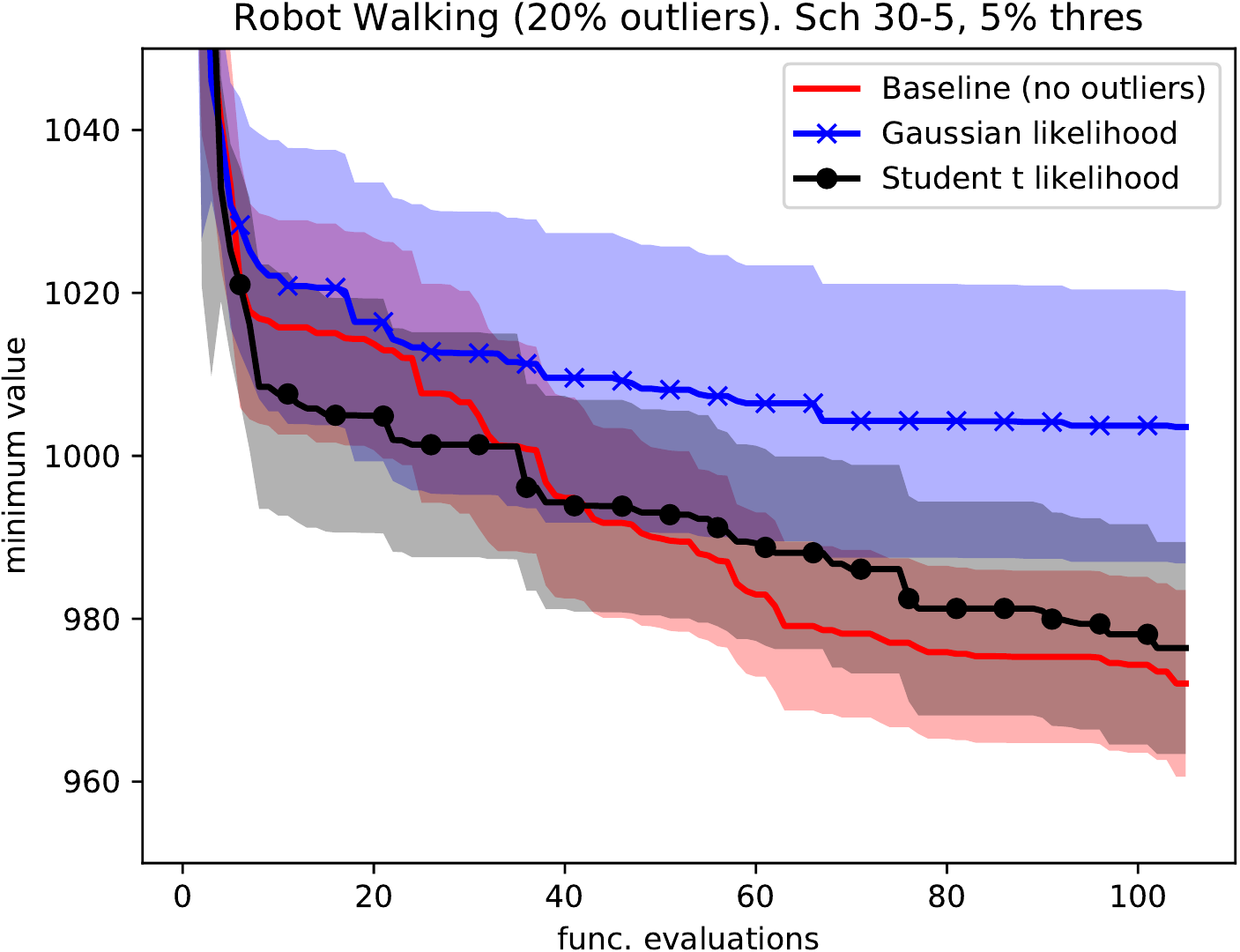}
    \hspace{1em}
    \includegraphics[width=0.4\linewidth]{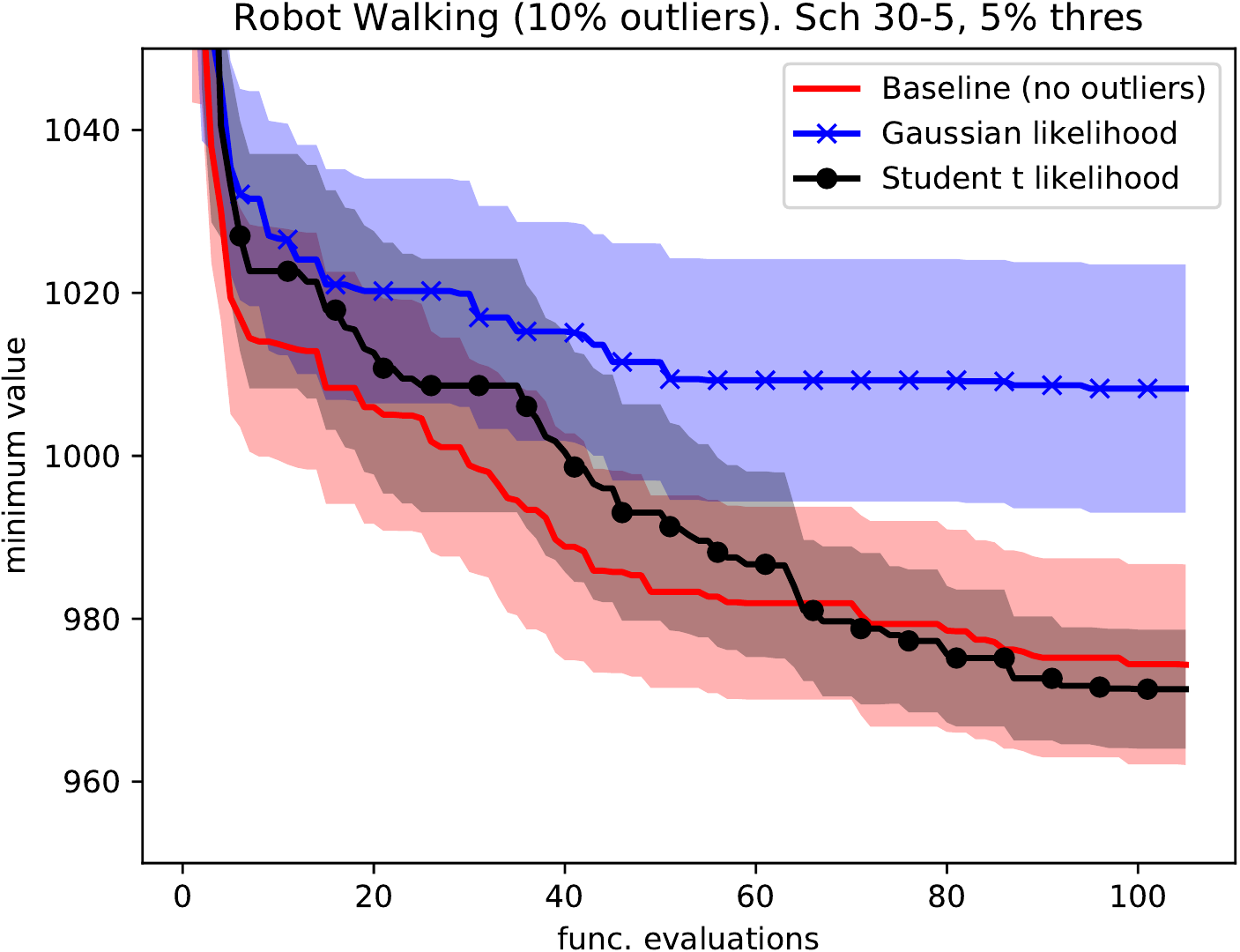}
  }
\end{figure}

\acks{We would like to thank Christopher G. Atkeson for releasing the
  code of the robot simulator and controller.}

\bibliography{robust}

\end{document}